\documentclass{article}

\usepackage{PRIMEarxiv}

\usepackage[utf8]{inputenc} 
\usepackage[T1]{fontenc}    
\usepackage{hyperref}       
\usepackage{url}            
\usepackage{booktabs}       
\usepackage{amsfonts}       
\usepackage{nicefrac}       
\usepackage{microtype}      
\usepackage{lipsum}
\usepackage{fancyhdr}       
\usepackage{graphicx}       
\graphicspath{{media/}}     
\usepackage{amsmath}
\usepackage{caption}

\usepackage{listings}
\usepackage{xcolor}

\lstdefinestyle{mypython}{
    language=Python,
    basicstyle=\ttfamily\small,      
    breaklines=true,                 
    frame=single,                    
    columns=fullflexible,
    keywordstyle=\color{blue},       
    commentstyle=\color{gray},       
    stringstyle=\color{brown},       
    captionpos=b                     
}

\title{Iti-Validator: A Guardrail Framework for Validating and Correcting LLM-Generated Itineraries
}

\author{
 Shravan Gadbail \\
  International Institute of Information Technology\\
  Hyderabad, India \\
  \texttt{shravan.gadbail@students.iiit.ac.in} \\
   \And
 Masumi Desai \\
  International Institute of Information Technology\\
  Hyderabad, India \\
  \texttt{masumi.desai@students.iiit.ac.in} \\
  \And
 Kamalakar Karlapalem \\
  International Institute of Information Technology\\
  Hyderabad, India \\
  \texttt{kamal@iiit.ac.in} \\
}

\begin{document}
\maketitle

\begin{abstract}
The rapid advancement of Large Language Models (LLMs) has enabled them to generate complex, multi-step plans and itineraries. However, these generated plans often lack temporal and spatial consistency, particularly in scenarios involving physical travel constraints. This research aims to study the temporal performance of different LLMs and presents a validation framework that evaluates and improves the temporal consistency of LLM-generated travel itineraries. The system employs multiple state-of-the-art LLMs to generate travel plans and validates them against real-world flight duration constraints using the AeroDataBox API. This work contributes to the understanding of LLM capabilities in handling complex temporal reasoning tasks like itinerary generation and provides a framework to rectify any temporal inconsistencies like overlapping journeys or unrealistic transit times in the itineraries generated by LLMs before the itinerary is given to the user. Our experiments reveal that while current LLMs frequently produce temporally inconsistent itineraries, these can be systematically and reliably corrected using our framework, enabling their practical deployment in large-scale travel planning.
\end{abstract}


\section{Introduction}
Recent advances in Large Language Models (LLMs) have enabled them to generate coherent and contextually relevant text across various domains, including travel planning. A notable use case is automated itinerary generation, where LLMs produce multi-step travel plans from user prompts. While these plans may appear well-structured, they often overlook real-world constraints.

Temporal consistency—ensuring realistic transit times, and avoiding overlapping events—is a key but often ignored aspect. Without grounding in such constraints, LLMs can generate infeasible itineraries, such as scheduling simultaneous activities in different cities or suggesting impossible travel durations.

To address this limitation, we propose Iti-Validator, a guardrail framework designed to evaluate and correct LLM-generated travel itineraries. The framework leverages real-world travel data, specifically flight duration obtained from the AeroDataBox API—to verify the feasibility of generated plans. By validating and, if necessary, correcting these itineraries before presenting them to the user, Iti-Validator ensures that generated outputs not only read well but also adhere to realistic temporal constraints. While this work focuses on itineraries involving commercial flights, this scope is intentional: the commercial aviation sector is a trillion-dollar industry, and the ability to automatically generate and validate such itineraries has significant real-world applicability. Airlines, travel agencies, and itinerary planning platforms can directly integrate such a system to improve operational efficiency and customer satisfaction.

\noindent This work makes the following contributions:
\begin{itemize}
    \item We conduct a systematic analysis of multiple state-of-the-art LLMs to assess their performance on temporal reasoning tasks within the domain of travel planning.
    \item We introduce a validation pipeline that integrates external knowledge sources (AeroDataBox for real-world travel feasibility) to detect and rectify temporal inconsistencies in generated itineraries.
\end{itemize}

\section{Related Work}
Recent work has explored the use of LLMs for travel itinerary generation, spatiotemporal reasoning, and path planning. \textbf{TripCraft}~\cite{Chaudhuri2025TripCraft} presents a benchmark for LLM-based travel planning under realistic constraints such as transfer windows and transit durations. Similarly, \textbf{TravelPlanner}~\cite{Xie2024TravelPlanner} introduces a planning challenge where language agents must generate valid itineraries under multi-step logical and temporal constraints, highlighting how even state-of-the-art LLMs struggle without explicit constraint enforcement. \textbf{ItiNera}~\cite{Tang2024ItiNera} integrates spatial clustering and optimization with LLMs to produce urban itineraries for real users, focusing on efficient routing within a city. \textbf{Geo-LLaMA}~\cite{Li2024GeoLlama} tackles trajectory generation by injecting geospatial constraints into LLM outputs. These works primarily focus on optimizing or learning better itinerary generation within the LLM itself.

In parallel, a line of research examines the spatial and temporal reasoning abilities of LLMs more generally. For spatial path planning, \textbf{LLM-A*}~\cite{Meng2024LLMAstar} combines LLM reasoning with the A* algorithm to ensure valid navigation routes. \textbf{PPNL}~\cite{Aghzal2023PathPlanning} introduces a benchmark for spatial-temporal reasoning in natural language, while \textbf{Zhang et al.}~\cite{Zhang2025S2ERS} mitigate spatial hallucinations using reinforcement learning-enhanced prompting. For temporal reasoning, \textbf{ToT}~\cite{Fatemi2024ToT} evaluates LLMs on logical time arithmetic tasks, and \textbf{TCP}~\cite{Ding2025TCP} benchmarks collaborative project planning with interdependent temporal constraints.

Despite these advances, prior work either enforces constraints during generation or uses learned models to improve output quality. \textbf{In contrast, our approach introduces a post-processing temporal validator that operates independently of the LLM.} Unlike methods such as \emph{ItiNera} or \emph{Geo-LLaMA}, our validator requires neither fine-tuning nor integration with the LLM’s internal architecture, making it model-agnostic. To the best of our knowledge, this is the first work to systematically validate and correct LLM-generated travel itineraries using real-world flight data accessed via an API (AeroDataBox) as an external knowledge source.

\section{Methodology: Iti-Validator Framework}
\begin{figure}[h]
  \centering
  \includegraphics[width=1\textwidth]{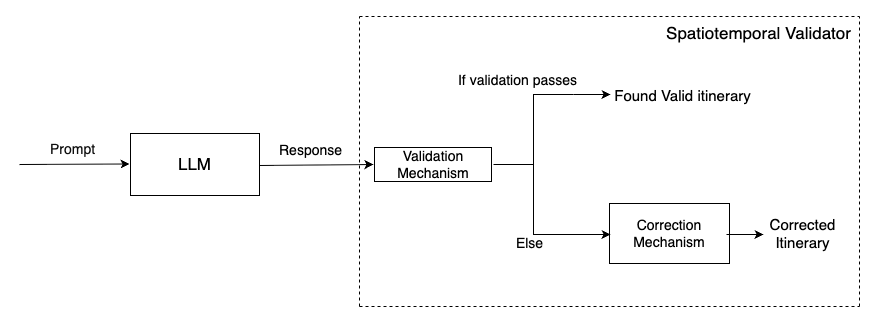}
  \caption{Iti-Validator}
  \label{fig:arch}
\end{figure}

The proposed system implements a multi-stage validation and correction framework for travel itinerary generation. The architecture consists of four main components: (1) an LLM-based itinerary generator, (2) the AeroDataBox API integration for real-world flight time retrieval, (3) a validation mechanism that checks itinerary consistency against temporal rules, and (4) a correction mechanism that adjusts the itinerary to fix any detected issues. Figure~\ref{fig:arch} presents an overview of the system.

\subsection{Validation Rules}
To evaluate an itinerary, the validator applies a set of rules which aim to account for real-world travel constraints:
\begin{itemize}
    \item \textbf{No Overlap Rule:} A traveler cannot be in two places at the same time. The itinerary is invalid if any visit intervals overlap or if a departure time from one city precedes the arrival time to that same city.
    \item \textbf{Minimum Transit Time Rule:} Traveling between two distinct cities requires a minimum non-zero time. We assume all inter-city travel is by commercial aircraft (the fastest conventional mode for long-distance travel). For each pair of consecutive destinations, the validator fetches the minimum flight duration $t_{minimum}$ between those cities using the AeroDataBox API. If the itinerary allocates less travel time than $t_{minimum}$ between a departure and the next arrival, that segment is marked invalid.
    \item \textbf{Maximum Transit Time Rule:} An itinerary leg should not contain excessively more time than needed. We set an upper bound that travel time should not exceed $2 \times t_{minimum}$ (double the minimum flight duration, to accommodate layovers or realistic scheduling buffers). If an LLM allocates an extremely large gap for a flight (e.g., several days for a 10-hour flight), that segment is also flagged as invalid.
    \item \textbf{Minimum Stay Rule:} Each destination should have a minimum stay duration (we use two days in our experiments, unless specified otherwise). This ensures the itinerary is realistic for a multi-city tour, avoiding itineraries where a traveler lands and departs immediately with no time spent.
\end{itemize}

\noindent The validator parses the LLM's JSON itinerary (which includes location names and timestamped arrival/departure for each stop) and checks each leg against these rules. If all segments satisfy the rules, the itinerary is labeled \emph{valid}. If any rule is violated, the itinerary is \emph{invalid} and needs correction.

\noindent While our current implementation assumes international travel by commercial flights, the validation rules are modular and easily extendable. For instance, adding trains or road travel would involve incorporating corresponding minimum and maximum transit times from relevant datasets. Similarly, domain-specific rules (e.g., budget limits, visa constraints, specific dates) can be layered onto the framework without altering its core structure.

\subsection{Correction Mechanism}
For each segment that fails validation, our framework automatically adjusts the schedule to fix the issue:
\begin{itemize}
    \item If a segment's allocated travel time is too short (violating the Minimum Transit Time Rule), we push the arrival time at the destination forward. Specifically, for a segment where city A's departure time is $t_{\text{dep}}^{A}$ and city B's arrival time is $t_{\text{arr}}^{B}$, if $t_{\text{arr}}^{B} - t_{\text{dep}}^{A} < t_{min}(A,B)$, we set
    \[
    t_{\text{arr}}^{B} := t_{\text{dep}}^{A} + t_{min}(A,B)\,,
    \]
    to ensure enough transit time.
    \item If a segment's travel time is excessively long ($t_{\text{arr}}^{B} - t_{\text{dep}}^{A} > 2 \times t_{min}$), we pull the arrival time earlier to a reasonable window. We cap it at $2 \times t_{min}$ after departure.
    \item If an overlap is detected (city B's arrival precedes city A's departure), we shift the subsequent schedule forward in time until no overlaps remain.
    \item If a city’s stay is shorter than the minimum, we extend the departure time from that city to meet the 2-day threshold.
\end{itemize}

After applying these modifications to all problematic segments, the itinerary is checked again to ensure all constraints are satisfied. In practice, a single pass of adjustments resolves all conflicts in our test cases.

We also explored an alternative correction approach: having the LLM itself regenerate the itinerary after being told what issues to fix. The validator can produce a corrective prompt listing each problem (e.g., “Flight from X to Y is too short, needs at least N hours; extend stay in Z to 2 days”) and ask the LLM to produce a new itinerary. \textit{We tried this iterative re-prompting (up to three attempts) for the three LLMs on 4-city itineraries. However, this approach proved unreliable—many itineraries remained invalid even after multiple retries, as the LLMs would fix some issues but introduce or leave others}. Given these inconsistent results, we discarded the self-correction approach in favor of deterministic rule-based adjustments, which guarantee a valid itinerary whenever one is feasible.

\subsection{Implementation Details}

\paragraph{LLM Itinerary Generation:} We experiment with three LLMs to generate itineraries, allowing us to compare their baseline performance and illustrate the need for validation. The models are: (1) OpenAI’s \textbf{GPT-4o-mini}, (2) Google’s \textbf{Gemini-2.0} (flash-thinking experimental preview), and (3) Meta’s \textbf{LLaMA2-7B} (7-billion-parameter open-source model). Each model is prompted to produce a multi-city trip itinerary in JSON format, listing each destination with arrival and departure timestamps (UTC, 24-hour format). We explicitly instruct a minimum two-day stay per city in the prompt to enforce the stay rule at generation time, though the models do not always comply.

\paragraph{AeroDataBox API Integration:} We use the AeroDataBox service to obtain realistic flight durations between city pairs. Given two cities (identified by major airport codes), the API returns typical flight times. The returned duration is parsed (hours and minutes) and converted to seconds for comparison. We add a fixed buffer (4 hours) to each flight duration as the minimum required travel time to account for airport logistics. If the API fails or no data is available for a route, we retry the API call, upto a maximum of 3 times. All API queries are cached to avoid redundant requests when validating multiple itineraries with common segments. In cases where the API fails repeatedly or returns null data, our framework proceeds only with non-null results to ensure that all validations are based on reliable flight duration information. This approach maintains the robustness of the system even when certain route data is unavailable.

The framework, implemented in Python, is designed to be flexible and can integrate with any LLM API. The validation and correction modules are lightweight rule-based scripts, and the average runtime per itinerary was under 3 seconds for 6-city trips including correction, excluding API latency for OpenAI's 4o-mini while it was under 6 seconds for gemini and under 10 seconds for llama2-7b. The overall process is:

\begin{enumerate}
    \item LLM generates itinerary.
    \item Validator checks each segment via AeroDataBox and the rules.
    \item If any segment violates a rule, adjust travel times.
    \item Output the corrected itinerary (or report it as valid if no changes needed).
\end{enumerate}

If an itinerary had format-related issues (e.g., JSON-formatting, date-time formatting), the framework would retry the generation of the itinerary by re-prompting, upto a maximum of three retries, the LLM by specifying the exact issue along with the base prompt. 

\begin{figure}[t]
\centering
\begin{minipage}{0.9\linewidth}
\begin{verbatim}
[
    {{
        "place": "city_name (IATA)",
        "arrival_time": "YYYY-MM-DD HH:MM",
        "departure_time": "YYYY-MM-DD HH:MM"
    }},
    {{
        "place": "city_name (IATA)",
        "arrival_time": "YYYY-MM-DD HH:MM",
        "departure_time": "YYYY-MM-DD HH:MM"
    }}
    // ... up to {num_destinations} items
]
\end{verbatim}
\end{minipage}
\caption{JSON schema used for LLM-generated itineraries. 
Each entry records the city, arrival timestamp, and departure timestamp. 
The full schema, code and detailed experiment results are available at \textcolor{blue}{
\url{https://github.com/MasumiD/Spatiotemporal-Validator-for-LLMs}.}}
\label{fig:json-schema}
\end{figure}

\subsection{Prompt Construction and Deployment}
\label{sec:prompt-construction}

A key aspect of our framework is how prompts are constructed and used during itinerary generation and correction. Below, we provide direct excerpts from the source code that illustrate each stage of the process.

\paragraph{1. Fixed Sequence Prompt:}
When a fixed route sequence is specified, the following prompt template is used.
\begin{lstlisting}[language=Python, caption={Fixed sequence prompt construction.}]
prompt = f"""
You are tasked with creating a valid time schedule for a PRE-DEFINED travel itinerary.
The itinerary visits {current_num_destinations} destinations.
You MUST follow this exact sequence of cities and use their IATA codes as provided:
{fixed_route_str_for_prompt}

The cities involved are from the following list (for context and ensuring correct naming/IATA):
{cities_str}

IMPORTANT: Return ONLY a valid JSON object with this EXACT structure (no additional text, no markdown formatting):
{{
    "itinerary": [
        // Example for the first stop, ensure "place" matches the fixed sequence
        {{
            "place": "{fixed_route_sequence[0]['place']} ({fixed_route_sequence[0]['iata']})", 
            "arrival_time": "YYYY-MM-DD HH:MM",
            "departure_time": "YYYY-MM-DD HH:MM"
        }}
        // ... and so on for all {current_num_destinations} cities in the fixed_route_sequence
    ]
}}

Requirements:
1. All times MUST be in UTC.
2. Use 24-hour format (e.g., 14:30, 00:00 for midnight) ONLY and EXACTLY MATCH the format '%Y-%m-%d %H:%M'.
3. The "place" field in your JSON for each stop MUST EXACTLY MATCH the city name and IATA code from the fixed sequence provided above. Do not alter the sequence or the cities.
4. Do NOT add any explanatory text or markdown formatting.
5. Ensure the JSON is properly formatted with correct commas and brackets.
6. Account for minimum flight times between cities (use realistic minimum flight durations).
7. For each city, the difference between its 'departure_time' and 'arrival_time' (i.e., the stay at that city) MUST be more than 2 days (> 49 hours).
8. For each segment, the difference between the 'departure_time' of the previous city and the 'arrival_time' of the next city MUST be equal to the minimum realistic flight time (plus a 1 hour buffer for airport procedures). Do NOT add extra days or hours to the travel time.
9. The stay duration and the travel duration are separate: do NOT add the 2-day minimum stay to the travel time. The 2-day minimum applies only to the time spent at each city.
10. Return ONLY the JSON object, nothing else.
"""
\end{lstlisting}

\paragraph{2. Generic Generation Prompt:}
If no fixed sequence is provided, a generic template is employed:
\begin{lstlisting}[language=Python, caption={Generic generation prompt construction.}]
# Code excerpt for generic prompt
prompt = f"""
Generate a travel itinerary visiting {num_destinations} destinations, exclusively using air travel.
You MUST use ONLY these cities for your itinerary:
{cities_str}

IMPORTANT: Return ONLY a valid JSON object with this EXACT structure (no additional text, no markdown formatting):
{{
    "itinerary": [
        {{
            "place": "city_name (IATA)",
            "arrival_time": "YYYY-MM-DD HH:MM",
            "departure_time": "YYYY-MM-DD HH:MM"
        }},
        {{
            "place": "city_name (IATA)",
            "arrival_time": "YYYY-MM-DD HH:MM",
            "departure_time": "YYYY-MM-DD HH:MM"
        }}
        // ... up to {num_destinations} items
    ]
}}

Requirements:
1. All times MUST be in UTC.
2. Use 24-hour format (e.g., 14:30, 00:00 for midnight).
3. Travel dates must be between {date_suggestion_start} and {date_suggestion_end}.
4. Include the IATA airport code for each city in parentheses.
5. Do NOT add any explanatory text or markdown formatting.
6. Ensure the JSON is properly formatted with correct commas and brackets.
7. For each city, the difference between its 'departure_time' and 'arrival_time' (i.e., the stay at that city) MUST be more than 2 days (> 48 hours).
8. For each segment, the difference between the 'departure_time' of the previous city and the 'arrival_time' of the next city MUST be equal to the minimum realistic flight time (plus a 1 hour buffer for airport procedures). Do NOT add extra days or hours to the travel time.
9. The stay duration and the travel duration are separate: do NOT add the 2-day minimum stay to the travel time. The 2-day minimum applies only to the time spent at each city.
10. Return ONLY the JSON object, nothing else.
"""
\end{lstlisting}

\paragraph{3. Prepending Feedback (Correction Scenario):}
When the generated itinerary is invalid or incorrectly formatted, feedback is prepended to the base prompt:
\begin{lstlisting}[language=Python, caption={Prepending feedback in correction scenarios.}]
prompt_to_use = feedback_message + base_prompt
\end{lstlisting}

a. JSON formatting error
\begin{lstlisting}[language=Python, caption={Correction scenario: JSON formatting error.}]
feedback_message = """The previous response was not a valid JSON object. Please ensure:
1. The response is a single, valid JSON object
2. The JSON has an \"itinerary\" array containing exactly 4 stops
3. Each stop has \"place\", \"arrival_time\", and \"departure_time\" fields
4. All times are in UTC and follow the format 'YYYY-MM-DD HH:MM'
5. Each place includes the IATA code in parentheses
Example format:
{
    \"itinerary\": [
        {
            \"place\": \"London (LHR)\",
            \"arrival_time\": \"2024-03-20 10:00\",
            \"departure_time\": \"2024-03-20 14:00\"
        }
    ]
}"""
\end{lstlisting}

b. Time formatting error
\begin{lstlisting}[language=Python, caption={Correction scenario: Time formatting error.}]
feedback_message = f"""Error in time format for {stop.get('place', 'unknown')}. Please ensure:
1. All times are in UTC and follow the EXACT format 'YYYY-MM-DD HH:MM'
2. Use 24-hour format (e.g., 14:30, 00:00 for midnight)
3. Include leading zeros (e.g., '01:05' not '1:5')
4. No timezone indicators or UTC suffix
Example: '2024-03-20 14:30'"""
\end{lstlisting}

c. Insufficient stops error
\begin{lstlisting}[language=Python, caption={Correction scenario: Insufficient stops error.}]
feedback_message = f"""Generated itinerary has insufficient stops. Please ensure:
1. The itinerary contains exactly {num_destinations} stops
2. Each stop has all required fields (place, arrival_time, departure_time)
3. Each place includes the IATA code in parentheses
4. All times are in UTC and follow the format 'YYYY-MM-DD HH:MM'
Example format:
{{
    \"itinerary\": [
        {{
            \"place\": \"London (LHR)\",
            \"arrival_time\": \"2024-03-20 10:00\",
            \"departure_time\": \"2024-03-20 14:00\"
        }},
        {{
            \"place\": \"Paris (CDG)\",
            \"arrival_time\": \"2024-03-20 16:00\",
            \"departure_time\": \"2024-03-21 10:00\"
        }}
    ]
}}"""
\end{lstlisting}

These excerpts demonstrate the exact mechanism of how prompts are defined, corrected when necessary, and ultimately deployed in the LLM call.

\section{Evaluation Setup and Metrics}
We evaluated the LLMs' raw itinerary outputs using multiple metrics. Each LLM was tasked with generating itineraries for trips with a varying number of destinations (we tested 4, 5, and 6 destination trips, which equates to 3, 4, and 5 flights per itinerary respectively). The cities were chosen to ensure both short and long international flights, covering a wide range of cases, with examples including routes across North America, Asia, Europe, and the Middle East.

For each combination of model and number of cities, we collected multiple itineraries (100 per combination). We then applied the validator to each itinerary. To quantify performance, we use:
\begin{itemize}
    \item \textbf{\% Invalid Itineraries (raw):} The percentage of LLM-generated itineraries that contained one or more temporal inconsistencies before any correction. This indicates how often the LLM failed to produce a fully feasible plan.
    \item \textbf{\% Invalid Segments (raw):} Out of all individual travel segments (flight legs) in the itineraries, the percentage that violated constraints (either too short or too long). This measures how frequently an LLM misjudges a single leg’s timing.
    \item \textbf{Avg Issues per Itinerary:} The average number of issues (invalid segments) detected per itinerary. An itinerary with multiple problematic legs would increase this count.
\end{itemize}

\textit{Since our validator corrects any detected issue, the post-validation itineraries were all consistent}. Thus, our evaluation focuses on the severity of issues in the raw output and how much correction was needed.

\section{Results and Analysis}

\begin{table}[h]
\small
\centering
\begin{tabular}{llccc}
\hline
\textbf{LLM Model} & \textbf{Cities} & \textbf{Invalid Itin.} & \textbf{Invalid Seg.} & \textbf{Avg Issues/Itn.} \\
\hline
\text{Gemini-2.0}$^{\dagger}$  & 4 & \textbf{48}\% & \textbf{21.00}\% & \textbf{0.63} \\
GPT-4o-mini & 4 & 97\% & 78.00\% & 2.34 \\
LLaMA2-7B   & 4 & 100\% & 92.33\% & 2.77 \\
\text{Gemini-2.0}$^{\dagger}$  & 5 & \textbf{68}\% & \textbf{27.75}\% & \textbf{1.11} \\
GPT-4o-mini & 5 & 100\% & 82.00\% & 3.28 \\
LLaMA2-7B   & 5 & 100\% & 93.69\% & 3.75 \\
\text{Gemini-2.0}$^{\dagger}$  & 6 & \textbf{76}\% & \textbf{29.60}\% & \textbf{1.48} \\
GPT-4o-mini & 6 & 99\% & 78.00\% & 3.90 \\
LLaMA2-7B   & 6 & 100\% & 91.60\% & 4.58 \\
\hline
\end{tabular}
\caption{Temporal inconsistencies in raw LLM-generated itineraries, before applying our validator. “Invalid Itin.” is the percentage of itineraries with any temporal inconsistencies. “Invalid Seg.” is the percentage of flight segments that violate constraints. “Avg Issues/Itn.” is the average number of issues (segments or stays) per itinerary.(($\dagger$) Refers to the Gemini-2.0 variant: gemini-2.0-flash-thinking-exp-01-21.
)}
\label{tab:resultsSummary}
\end{table}
Table~\ref{tab:resultsSummary} summarizes the consistency of raw itineraries produced by each model, and how it varies with the number of destinations.

\noindent Our validator corrects 100\% of the invalid itineraries as it is a deterministic, rule-based validator. We check our LLMs for 100 itineraries each for 4,5 and 6 destination itineraries.
We observe that all models exhibit significant temporal inconsistency, with a clear increase in errors as itinerary complexity grows (i.e., more cities involved). Specifically, LLaMA2-7B consistently performs the worst, producing temporally invalid itineraries in 100\% of cases across all configurations, with over 90\% of flight segments violating constraints. The average number of temporal issues per itinerary also peaks for LLaMA2-7B, reaching 4.58 issues for 6-city plans. Note that the number of total segments for an itinerary is equal to one less than the number of destinations. So an itinerary with 6 destinations will have a total of 5 segments.

GPT-4o-mini shows slightly better performance but still suffers from high error rates. It generates invalid itineraries in over 97\% of cases for 4-city plans, increasing to 99–100\% for more complex scenarios. The average number of issues per itinerary reaches 3.90 for 6 cities, highlighting the model’s limited ability to handle real-world travel constraints.

On the other hand, Gemini-2.0 (Flash Variant) performs relatively better, particularly for smaller itineraries. With only 48\% invalid itineraries for 4-city plans and 1.48 average issues for 6 cities, Gemini demonstrates better temporal reasoning. However, the performance still degrades with more complex itineraries, suggesting that no model is inherently grounded in physical travel constraints.

\noindent Examining the types of errors, we observed distinct failure modes for different models:

\begin{itemize}
    \item \textbf{Underestimating Travel Time (Too Short):} GPT-4o-mini and Gemini commonly allocated far too little time for certain flights, leading to infeasible itineraries. GPT-4o-mini proposed a 4-city itinerary: New York $\rightarrow$ Tokyo $\rightarrow$ Dubai $\rightarrow$ Toronto. It scheduled an 8-hour window for the Tokyo–Dubai flight (which is too short by about 3 hours). In Table~\ref{tab:resultsSummary}, the fact that invalid segments percentages for gpt-4o-mini are high but not 100\% indicates some segments (especially shorter flights or same-continent destinations) were timed reasonably, but at least one long-haul flight in each itinerary was often under-timed.
    \item \textbf{Overestimating Travel Time (Too Long):} In contrast, LLaMA2-7B tended to err by overestimating—sometimes drastically. It would insert multi-day gaps for flights that take less than a day. In one LLaMA2-generated 4-stop trip, a flight from Paris to Tokyo was scheduled with a 5-day window. This obviously avoids overlaps, but it violates our maximum transit rule. Every LLaMA2 itinerary had at least one excessive travel time violation. Thus, LLaMA2 achieved 0\% feasible itineraries due to the opposite reason as GPT-4o-mini – it overshot instead of undershot.
    \item \textbf{Minimum Stay Violations:} Despite prompting for a 2-day stay, we found a few cases where the model gave a 1-day or 1.5-day stay for a city. For instance, one Gemini itinerary had the traveler arrive in Dubai on June 7 at 14:30 and depart June 9 at 08:00, which is about 1.7 days. Our validator caught these as well. They were less common than the flight timing issues, but did occur in some itineraries.
\end{itemize}

\noindent To illustrate the range of temporal inconsistencies encountered, Figure~\ref{fig:example-itinerary-errors} presents a sample invalid itinerary.
\begin{figure}[t]
\centering
\begin{minipage}{0.4\linewidth}
\small   
\begin{verbatim}
Sydney (SYD)
  Arrive: 2025-06-07 10:00 UTC
  Depart: 2025-06-08 06:00 UTC    
[Minimum stay violation (only 20h stay)]

Frankfurt (FRA)
  Arrive: 2025-06-08 18:00 UTC
  Depart: 2025-06-11 04:00 UTC    
[Too short travel time (SYD→FRA needs 
~21h, only 12h gap)]

Cairo (CAI)
  Arrive: 2025-06-15 12:00 UTC
  Depart: 2025-06-17 13:00 UTC    
[Too long travel time (FRA→CAI needs 
~5h, allocated ~4 days to travel)]

Casablanca (CMN)
  Arrive: 2025-06-17 22:00 UTC
  Depart: 2025-06-21 09:00 UTC
\end{verbatim}
\end{minipage}
\caption{Example of validation failures in an LLM-generated itinerary. 
This single example illustrates all types of errors. Note that the comments are illustrative and not part of the LLM’s raw output.}
\label{fig:example-itinerary-errors}
\end{figure}

\begin{table}[t]
  \caption{Before and after validation of the sample itinerary using Iti-Validator.}
  \centering
  \small
  \begin{tabular}{p{3cm}p{6cm}p{6cm}}
    \toprule
    \textbf{Segment} & \textbf{Before Validation} & \textbf{After Validation} \\
    \midrule
    Sydney (SYD) &
    Arrive: 2025-06-07 10:00 \newline Depart: 2025-06-08 06:00 \newline [Minimum stay violation (20h)] &
    Arrive: 2025-06-07 10:00 \newline Depart: 2025-06-09 10:00 \newline [2-day stay enforced] \\
    
    \addlinespace[2pt]
    Frankfurt (FRA) &
    Arrive: 2025-06-08 18:00 \newline Depart: 2025-06-11 04:00 \newline [Too short transit from SYD (12h vs req. 21h)] &
    Arrive: 2025-06-10 07:00 \newline Depart: 2025-06-12 07:00 \newline [Transit fixed, 2-day stay enforced] \\
    
    \addlinespace[2pt]
    Cairo (CAI) &
    Arrive: 2025-06-15 12:00 \newline Depart: 2025-06-17 13:00 \newline [Too long FRA$\rightarrow$CAI transit] &
    Arrive: 2025-06-12 12:00 \newline Depart: 2025-06-17 13:00 \newline [5h transit] \\
    
    \addlinespace[2pt]
    Casablanca (CMN) &
    Arrive: 2025-06-17 22:00 \newline Depart: 2025-06-21 09:00 &
    Arrive: 2025-06-17 22:00 \newline Depart: 2025-06-21 09:00 \\
    \bottomrule
  \end{tabular}
  \label{tab:before-after}
\end{table}

\noindent After applying our validator, \textbf{all the above issues were eliminated} in the final itineraries. In effect, the “Invalid Itin.” percentages drop to 0\% with validation, since every itinerary we could adjust became feasible. The validator had a 100\% success rate in fixing timing conflicts among those detected as it is a rule-based validator. In a handful of cases, an itinerary had minor format issues (e.g., missing an arrival time field); those were regenerated by the LLM with feedback prompts and these issues got resolved everytime mostly in the first retry.

Another important observation from our experiments is that LLMs lack accurate knowledge of real-world flight options. We attempted to prompt the LLMs to suggest actual flight itineraries (with airline names or realistic routes) to see if that might improve their timing estimates. Instead, the models suggested flights that do not exist. For e.g., one model gave a direct flight from Surat, India to Doha, Qatar, even though Surat’s airport has very limited international service. This suggests that the LLMs do not have reliable internal data on flight networks or schedules.

These results underscore the need for external validation mechanisms like Iti-Validator. Without such guardrails, even state-of-the-art LLMs are prone to generating itineraries that are temporally infeasible, reducing their utility in practical travel planning scenarios. The performance of the selected LLMs suggests that they were not trained on diverse flight data and struggle to generalize effectively.

By integrating our validator with LLMs, airlines can refine their flight plans to connect cities more practically, leading to more efficient travel. This benefits not only the users, but also the airlines.

\section{Conclusion and Future Work}
We introduced \textbf{Iti-Validator}, a guardrail framework to verify and correct travel itineraries generated by Large Language Models (LLMs). By cross-checking LLM outputs with real-world flight durations and enforcing simple temporal rules, our system transforms often-impractical AI-generated travel plans into feasible itineraries. Through experiments on itineraries from multiple LLMs, we demonstrated that temporal inconsistencies are frequent in raw outputs, and that our validator is effective in detecting and eliminating these issues. This post-processing approach significantly enhances the reliability of LLM-based travel planning without requiring any changes to the LLMs themselves.

From an efficiency perspective, the guardrail operates within a few seconds per itinerary, dominated by API calls and correction logic. All corrections are deterministic and scale linearly with the number of itinerary legs. Since the AeroDataBox API calls are cached, repeated city pairs incur negligible overhead. Prompting costs for LLM regeneration (in cases of formatting errors) were minimal in our experiments, as most retries succeeded on the first attempt. These results suggest that Iti-Validator is both scalable and cost-effective for real-world deployment.
    
This work is a step toward safer and more trustworthy AI planning tools. In future work, we plan to extend the validator to handle a wider range of scenarios—such as multi-modal transport (integrating trains or driving legs) and more complex user preferences or constraints (e.g., maximum budget, specific layover cities). Another direction is to develop the validator as a more interactive assistant: it could provide explanations for its corrections or integrate with the LLM in a dialogue (beyond one-shot prompts) to iteratively refine itineraries with the user in the loop. Lastly, while our rule-based method outperformed autonomous LLM re-prompting for corrections, a hybrid approach could be explored: using learning to predict when a simple time-shift fix is insufficient and a larger itinerary restructuring is needed. We anticipate that our guardrail concept can be applied broadly to other domains where LLMs generate structured plans that must adhere to real-world constraints.

\bibliographystyle{unsrt}

\end{document}